\documentclass[a4paper,conference]{IEEE/IEEEtran}

\usepackage{url}
\usepackage{graphicx} 
\usepackage{cleveref} 
\usepackage{subcaption} 
\usepackage{booktabs} 
\usepackage{amssymb} 
\usepackage[noadjust]{cite}

\begin{document}
\title{Borrowing from yourself: Faster future video segmentation
with partial channel update
}

\author{
  \IEEEauthorblockN{Evann Courdier}
  \IEEEauthorblockA{Idiap Research Institute\\EPFL\\
  Email: evann.courdier@idiap.ch}
  \and
  \IEEEauthorblockN{Fran{\c c}ois Fleuret}
  \IEEEauthorblockA{University of Geneva\\Idiap Research Institute\\
  Email: francois.fleuret@unige.ch}
}


\maketitle

\begin{abstract}
Semantic segmentation is a well-addressed topic in the computer vision literature, but the design of fast and accurate video processing networks remains challenging. In addition, to run on embedded hardware, computer vision models often have to make compromises on accuracy to run at the required speed, so that a latency/accuracy trade-off is usually at the heart of these real-time systems' design.
For the specific case of videos, models have the additional possibility to make use of computations made for previous frames to mitigate the accuracy loss while being real-time.

In this work, we propose to tackle the task of fast future video segmentation prediction through the use of convolutional layers with time-dependent channel masking.
This technique only updates a chosen subset of the feature maps at each time-step, bringing simultaneously less computation and latency, and allowing the network to leverage previously computed features.
We apply this technique to several fast architectures and experimentally confirm its benefits for the future prediction subtask.
\end{abstract}

\IEEEpeerreviewmaketitle

\section{Introduction}
As a central task in computer vision, image semantic segmentation is now a mature topic for which there exist very performant convolutional, and transformer networks.
However, the more specific video semantic segmentation task has been less addressed, while many applications involving embedded systems that run real-time would benefit from it,
such as autonomous driving or computer assisted surgery systems.

Tackling a complex real-time task often comes with a latency/accuracy trade-off,
where one usually loses some accuracy to match the real-time requirement.
Yet, properly leveraging the inherent sequential nature of videos may allow substantial time gain by reusing previous computation while also bringing relevant insight from past frames, as did \cite{shelhamer_clockwork_2016,li2018low} which run parts of the network conditionally to how much change there was from the previous frame.
In our setting, we find information from past frames to be particularly relevant when predicting the segmentation of a future frame.

In this work, we propose to leverage computation from previous frames and to reduce FLOPs by using channel-wise masked convolutions.
These convolutions process a full input as standard convolutions but compute only a subset of their output channels at a given time-step, following a pre-defined masking schedule. Previous works proposed to reduce computation by dropping part of the image \cite{verelst2020dynamic} depending on what changed \cite{skipconv}, we propose instead to drop part of the convolution kernel.

While the gain in FLOPs is quite evident as there are strictly fewer computation operations, the latency gain is not.
This is because our proposed masked convolutions perform two tensor indexing operations, while these can be very time-consuming on GPU hardware and could offset the time saved in the convolutions.
Therefore, care has to be taken when designing the channel mask so that the whole model still has lower latency.

Our proposed convolution design is specifically relevant for future prediction tasks dealing with a stream of correlated inputs, such as video streams.
It indeed allows every convolutional layer to have access to part of the previous timestep's output to make its prediction.

We apply this simple idea to existing segmentation networks to save FLOPs and wall-clock time.
Particularly, it limits the performance drop for future prediction compared to original networks which have just been ``slimmed'' to run faster, but that cannot leverage the sequential nature of videos.
We also apply channel-wise masked convolutions to slimmed networks to take advantage of both techniques.

Our contributions are:
\begin{itemize}
    \item We introduce a time-dependent channel-wise masking scheme for convolutional layers (\cref{model_section}).
    \item We apply our technique to three segmentation networks, and we adapt the training and evaluation procedure (\cref{expsetup_section}).
    \item We perform several experiments using these layers to show their benefits and limits (\cref{results_section}) and make our full code available\footnote{\url{https://github.com/theevann/fast-cwm-segmentation}}.
\end{itemize}

\section{Method}
\label{model_section}

This section motivates our main idea and describes our proposed convolutional layer and its masking scheme.

\subsection{Idea}

Our goal is to speed up video segmentation networks.
As we specifically deal with videos, consecutive frames are alike and lead to similar computed features.
We propose leveraging this temporal correlation by only updating a subset of the output channels of convolutions at each time-step and re-using the features of the previous frame for the rest of the channels that are not computed.

We introduce a ``channel-wise masked'' convolution (\cref{mod:mask_conv_layer}) that works like a normal one but uses a binary mask to select which output channels to compute.
This convolution has lower FLOPs by design, but we find that we have to put constraints on the masks (\cref{mod:mask_constraints}) to also have lower latency.
Additionally, we choose to use a predefined finite set of masks and define a generator that will create these masks (\cref{mod:mask_generation}).

Our proposed channel-wise masked (CWM) convolutions can be used in place of the normal ones in any convolutional model to make it faster. We will denote a model using CWM convolutions a CWM-model.

\subsection{Masked convolutional layers}
\label{mod:mask_conv_layer}

\begin{figure}
  \centering
  \includegraphics[width=3.3in]{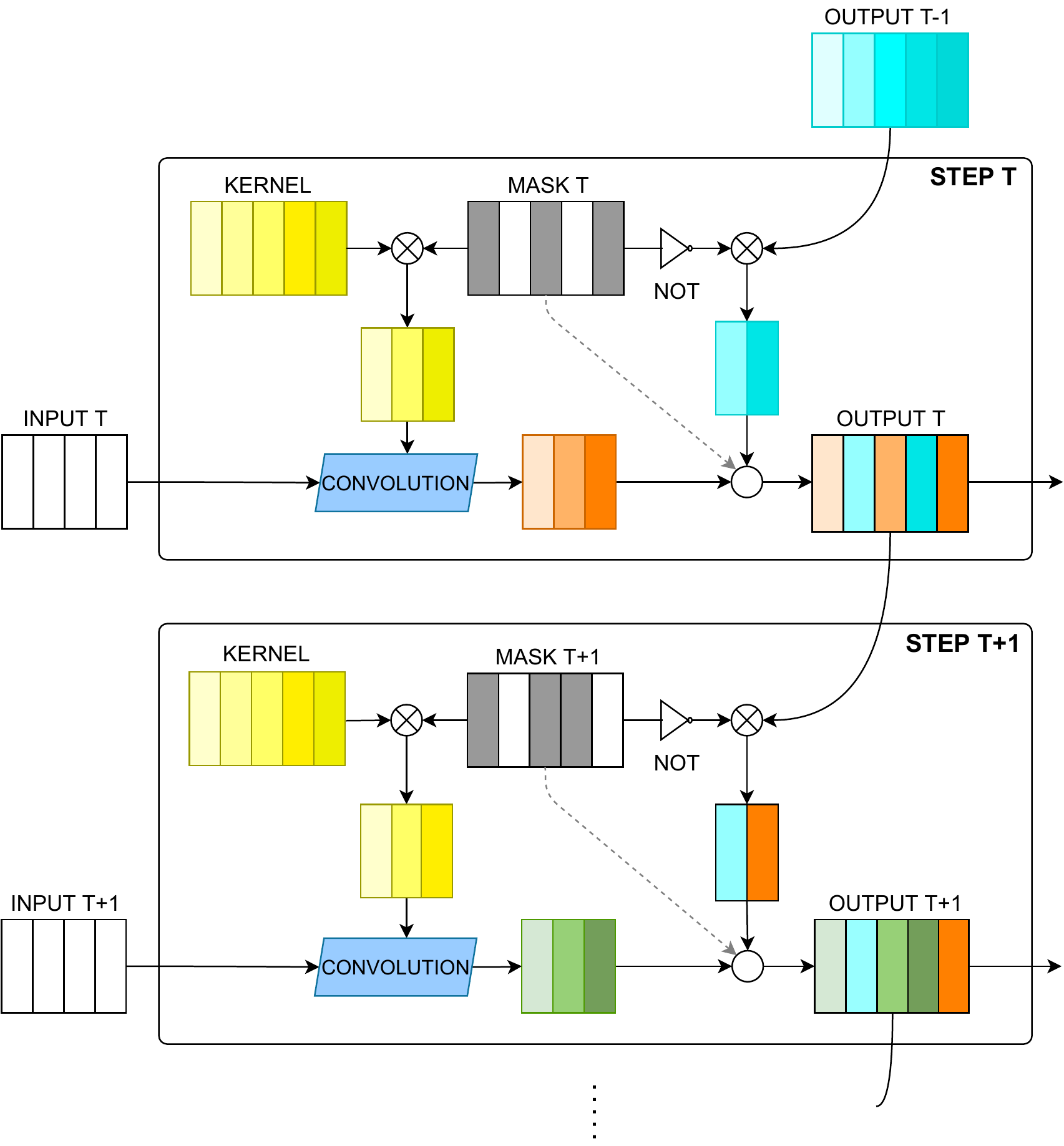}
  \caption{Operations performed in a channel-wise masked convolution. The active channels in the mask are greyed. The crossed circle is a masked select. The blank circle represents the interlacing of the current and previous output following the mask indexing information represented by the dashed line.
  }
  \label{fig:mask_conv}
  \vskip -.1in
\end{figure}

A channel-wise masked (CWM) convolution is a convolutional layer that uses a binary mask at each forward pass to select which output channels are computed.
In practice, this boils down to indexing the kernel tensor with the mask before performing the computation.
The result of this convolution is then combined with the previous time-step's result by simply replacing the older features with fresh ones depending on the mask, as can be seen on \cref{fig:mask_conv}.

This design implies that we have to save the output of every convolution for the next time step.
At the very first time-step though, there is no previous output to use, so we perform a full convolution without masking.
We noticed this is effective to properly initialize the saved outputs.

\subsection{Constraints on binary masks}
\label{mod:mask_constraints}

Depending on the time-step, our masked convolutions perform different computations. All input channels are still used, but each time-step sees a different group of convolution's output channels computed, specified through the use of binary masks. However, improper design of these masks
can lead to a significant latency increase, and we therefore have to put constraints on their structure and generation.

\textbf{Pre-defined}
While these masks could be chosen online, our initial experiments showed that doing so comes with a significant increase in processing time without accuracy gain. Therefore, for the rest of the paper, we only use a chosen number of pre-defined masks picked ahead of time. The masks are used sequentially, and once all masks are exhausted we restart from the first mask.


\textbf{Same number of channels} When two masks have a different number of active channels, CWM convolutions using those masks also have different latency. 
Since our main goal is to reduce maximum latency,
we need each time-step to have a similar computation time, so we have to activate the same number of channels in every mask.

\textbf{Contiguous}
Dealing with latency is deceptively complex, especially when working with GPU hardware.
In particular, non-contiguous tensor indexing on GPUs can be quite slow, and in our case, two indexing operations are required first to extract the convolution kernel and then to copy the result into the previous output.
While this aspect may seem to be a detail, we empirically observed that it matters significantly. In fact, using non-contiguous masks in some initial experiments increased the latency up to 15\% of its initial value.
Therefore, to realize the wall clock benefits of using fewer convolutional kernels (via masking), we enforce the masks to be contiguous.

\subsection{Mask generation}
\label{mod:mask_generation}

The mask generation process has to respect the constraints mentioned above and creates a set of masks that are {contiguous} and have the {same number of active channels}.
We delegate the generation process to a ``generator'' object that creates the set of masks and, given a time-step, returns the corresponding mask.

In the rest of the paper, we will use a specific generator called \textbf{bi-step generator (BG)} that creates only two different masks for two different time-steps. The first mask has a proportion of channels activated from the start, while the second mask has the same proportion of channels activated from the end. This proportion is at least equal to 50\%, and can be higher, which creates an overlap that corresponds to channels that are always activated.
Using contiguous masks in this manner makes it possible for some channels to be activated at every step while others will be activated at every other step.

\begin{figure}
  \centering
  \includegraphics[width=0.8\linewidth]{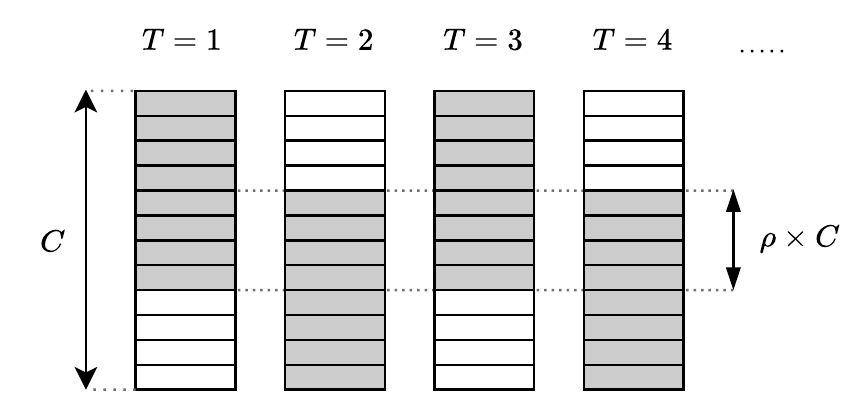}
  \caption{Masks provided for the 4 first time-steps in a bi-step generator (BG) for a convolution with $C$ output channels. Active channels are greyed. There are only two distinct masks used every other step. The central part which is always active has $\rho \times C$ channels. The generator is denoted $\rho$-BG.}
  \label{fig:bg}
  \vskip -.1in
\end{figure}

We will denote this type of mask generator using the notation $\rho$-BG, where $\rho \in [0,1]$ is a ratio that corresponds to the proportion of always active channels, as shown in fig \ref{fig:bg}. Therefore, a $1$-BG will make two full-ones masks (all channels always active), and a $0$-BG will make two half-full masks (no commonly active channel).

This bi-step generator design is interesting for two reasons. First, it keeps features relatively ``up to date'' as any output channels is recomputed at least every other step.
Second, the kernel weights used every other step can learn to distinguish recent from old input channels.
Other generators with more steps and masks can be used, but respecting the constraints greatly reduces the set of usable masks.
In the supplementary, we compared performance between a bi-step generator and a random contiguous mask generator.

\section{Experimental Setup and Models}
\label{expsetup_section}

\subsection{Dataset}

The Cityscapes dataset \cite{cordts2016cityscapes} is one of the few dense pixel-level scene semantic segmentation datasets.
We picked it as it provides a high number of video sequences and not only isolated frames.
Each sequence contains $30$ frames, and the $20^{th}$ frame is annotated with fine pixel-level class labels for $19$ object categories.
In total, it contains $2975$ training, $500$ validation and $1525$ testing video sequences.

During training, we use image crops of $700 \times 700$.
We perform standard image augmentations with random horizontal flip, random scaling from $0.75$ to $1.5$, and random Gaussian blur.
We apply the same augmentation to all images in a sequence.

Our focus is on future segmentation prediction, \textit{i.e.,}  networks are trained to predict a segmentation at $T+1$ from an image at time $T$.
As we use Cityscapes, we take the $19^{th}$ frame of the sequence as input, and the network is trained to predict the $20^{th}$ frame segmentation (for which we have ground-truth).

\subsection{Networks}
\label{exp:networks}

For our experiments, we apply our channel-wise masked (CWM) convolution to three different networks. We replace every convolution layer with its CWM counterpart, except for the very first and very last convolution, as well as convolutions used for skip connection (when block's input and output have different sizes).
We use:
\begin{itemize}
  \item \textbf{SwiftNet} \cite{orsic2019defense}, a performant light-weight network for real-time segmentation, with a ResNet-18 \cite{he2016deep} backbone.
  We train it using the Adam optimizer with default parameters. We use an initial learning rate of $4e-5$ and a weight decay of $1e-5$, along with a cosine annealing lr-schedule with $\eta_{min} = 1e-6$.

  \item \textbf{SFNet} \cite{li2020semantic}, the state-of-the-art for real-time segmentation network on Cityscapes, with a ResNet-18 backbone.
  We train it using the SGD optimizer with a momentum of $0.9$, an initial learning rate of $5e-3$ and a weight decay of $5e-4$. We use a poly lr-schedule with a power of $0.9$.

  \item \textbf{DeepLab V3+} \cite{chen2018encoder}, an accurate deep segmentation network, with a ResNet-18 backbone.
  We train it using the SGD optimizer with a momentum of $0.9$, an initial learning rate of $5e-2$ and a weight decay of $5e-4$. We use a poly lr-schedule with a power of $3$.
\end{itemize}

All hyperparameters mentioned above are taken from respective papers, and we train all networks for $500$ epochs with batch-size $6$.
However, we lower original learning rates since we initialize models with pretrained weights from original networks trained to predict the segmentation of the current input.

Unless otherwise specified, we use a $\rho$-BG (defined in \ref{mod:mask_generation}) as mask generator in our experiments.
Specifically, we conduct most of our experiments using two generators: $0$-BG and $0.25$-BG.

\subsection{Slimming networks}
\label{exp:slim}

Replacing convolutions with their CWM version reduces FLOPs and latency, but hurts accuracy.
To assess the advantage in using CWM convolutions, we use as baseline the original network, which has been slimmed in the way presented in \cite{howard2017mobilenets}.

This technique consists in thinning a network uniformly at each layer by choosing a width multiplier $\alpha \in  (0, 1]$ and multiplying the number of input and output channels in every convolution by $\alpha$.
Original networks are slimmed ahead of time and then trained normally.
Such slimming reduces the FLOPs, the latency, and the number of parameters, and hence allows comparing slimmed-network - our baseline - and CWM-network performance at the same computational cost.

In addition, we note that slimming is a simple complementary approach to our method to trade off computation for performance. Therefore, our main experiments combine the two techniques, using the four width-multipliers $\alpha \in \{0.5, 0.65, 0.8, 1\}$.


\subsection{Evaluation}
\label{exp:eval}

As specified in section \ref{mod:mask_conv_layer}, our CWM models perform normal (full) convolutions at the first time step. Therefore, we expect the model's behavior in the first few steps to be slightly different from its steady-state's behavior, \textit{i.e.} after the model has processed multiple images.


To have a correct estimation of the model's steady-state performance, we introduce Asymptotic Behavior Testing (ABT), which consists in feeding the model with inputs from $T-k$ up to $T-1$, with $k$ high enough for the network to run in steady-state.
In our case, we use the highest possible value $k=19$ permitted by this dataset.
Note that ABT is equivalent to a normal evaluation for original models, which do not use the sequential nature of the inputs.

Moreover, our CWM models have a different behavior every other step. Thus, to have a fair estimation of its expected asymptotic performance, we average the mIoU obtained with ABT using $k=19$ and $k=18$ in all experiments. 


The performance metric used in all our experiments is the mean Intersection over Union (mIoU), computed on Cityscapes' validation set.
All timing measurements are done on a GTX1080 GPU.

\subsection{Training details}
\label{exp:training}


\textbf{Length of the input sequence}
As for evaluation, we also train our model with a sequence of images starting at offset $j$, which practically means that we feed the model with inputs from $T-j$ to $T-1$, and use the final prediction for optimization.
We experimentally set this offset to $j=9$ for our main experiment (\cref{res:main}) and unless otherwise specified we use $j=7$ for our additional experiments, a trade-off between training time and testing performance. 


\textbf{Bi-sequence training}
Since our models have a different behavior every other step, we propose training our models with two input sequences starting at consecutive offsets. Concretely, for each image sequence, we first process and perform an optimization step with the subsequence \{$T-j$\,\,\dots\,\,$T-1$\}, and then again with the subsequence \{$T-j+1$\,\,\dots\,\,$T-1$\}.



\section{Results}
\label{results_section}

In this section, we first evaluate the performance of CWM models compared to original models for future segmentation prediction.
Then, we study the relevance and influence of various design choices and hyperparameters.
In all experiments, we compute the mIoU with asymptotic behavior testing (presented in \ref{exp:eval}) on Cityscapes' validation set.
We mostly included here plots of wall-clock times, additional plots of FLOPs are in Appendix.

\subsection{Future segmentation prediction}
\label{res:main}

\begin{figure*}
  \centering
  \includegraphics[width=.9\linewidth]{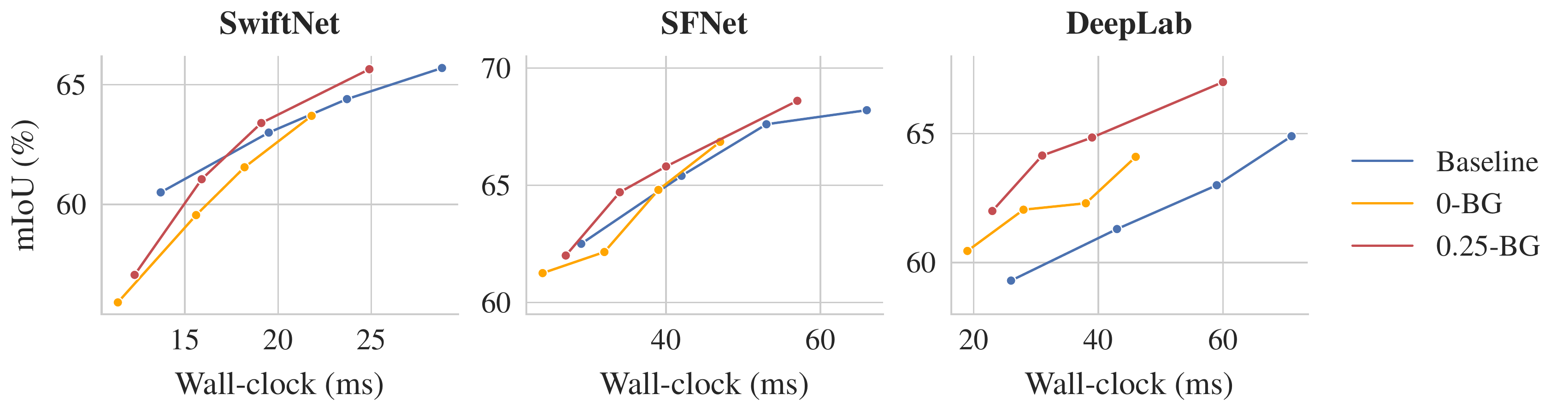}
  \includegraphics[width=.9\linewidth]{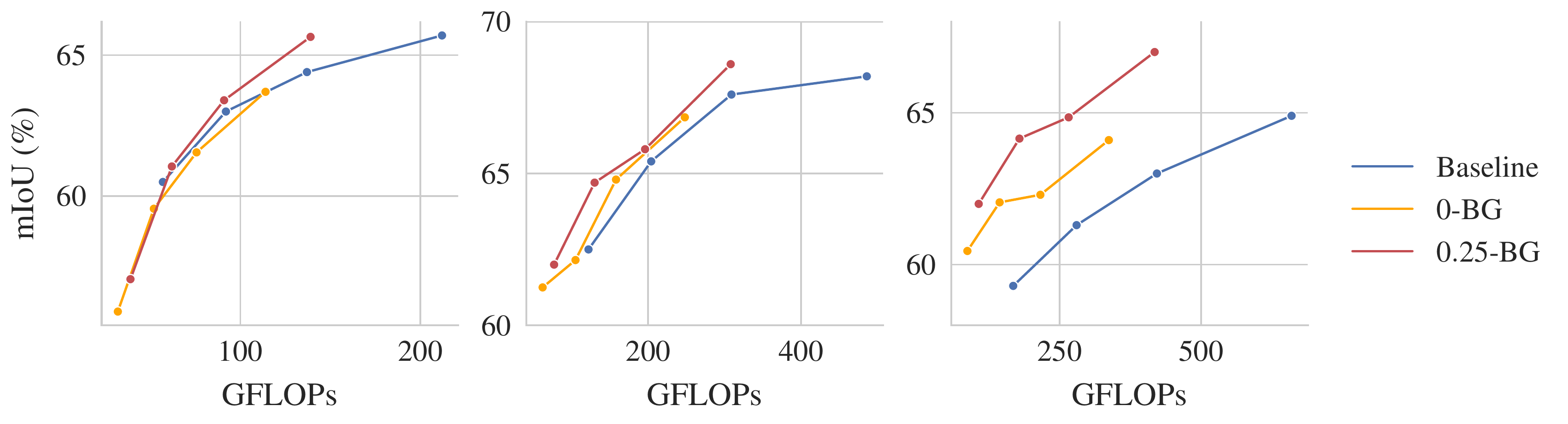}

  \caption{mIoU for three different models with and without CWM convolutions. Baseline is the original model, $0$-BG and $0.25$-BG are our CWM versions of the model using bi-step generators defined in \cref{mod:mask_generation}.
  Each line corresponds to a model for which the speed/accuracy trade-off is modulated with the width multipliers $\alpha \in \{0.5, 0.65, 0.8, 1\}$ to slim the network (\cref{exp:slim}).
  Our un-slimmed models using 0.25-BG, corresponding to rightmost red points, all have a better mIoU and a lower latency than original models.}
  \label{fig:main}
  \vskip -0.05in
\end{figure*}

In figure \ref{fig:main}, we compare baseline models to our CWM versions of those with generators $\rho$-BG for $\rho \in \{ 0, 0.25 \}$. Each connected line corresponds to a model for which the trade-off between computation and accuracy is controlled by changing the width multipliers $\alpha \in \{0.5, 0.65, 0.8, 1\}$ (\cref{exp:slim}).


For every line, the rightmost points correspond to the un-slimmed version of the model.
On all figures, un-slimmed 0.25-BG models not only use as expected less time and flops, but also have a higher mIoU than base models.
This is likely because the layers of our CWM-models have access to part of previous feature maps.
Having access to both previous and current features allows to more accurately estimate speeds and predict the future position of a moving object.

As we slim more and more with a lower $\alpha$, the number of active channels per time-step gets too small to make proper predictions.
This is especially true for SwiftNet and SFNet that are already both very optimized networks.
With these models, we see that using CWM convolutions is better than using original ones when networks are slightly slimmed. In particular, using un-slimmed CWM models with 0.25-BG decreases wall-clock by 15\% and FLOPs by 35\% for the same mIoU.
When slimming these networks more, we then get quite comparable performances as using CWM convolutions.

For DeepLab, using CWM convolutions leads to much more competitive results. In particular, we see that a slimmed model using a 0.25-BG can reach a similar mIoU as the base model with a decrease of 45\% wall-clock and 60\% FLOPs. Moreover, the un-slimmed 0.25-BG model (rightmost red dot) has about 2\% higher mIoU than the base model while being 15\% faster.

Generally, we can see that using CWM-convolutions is a better way to accelerate a model than a simple slimming.
These results demonstrate a real potential for CWM convolutions to speed up future prediction networks.

\subsection{Controlling the speed/accuracy trade-off}
\label{res:masks}

\begin{figure}
  \centering
  \includegraphics[width=0.9\linewidth]{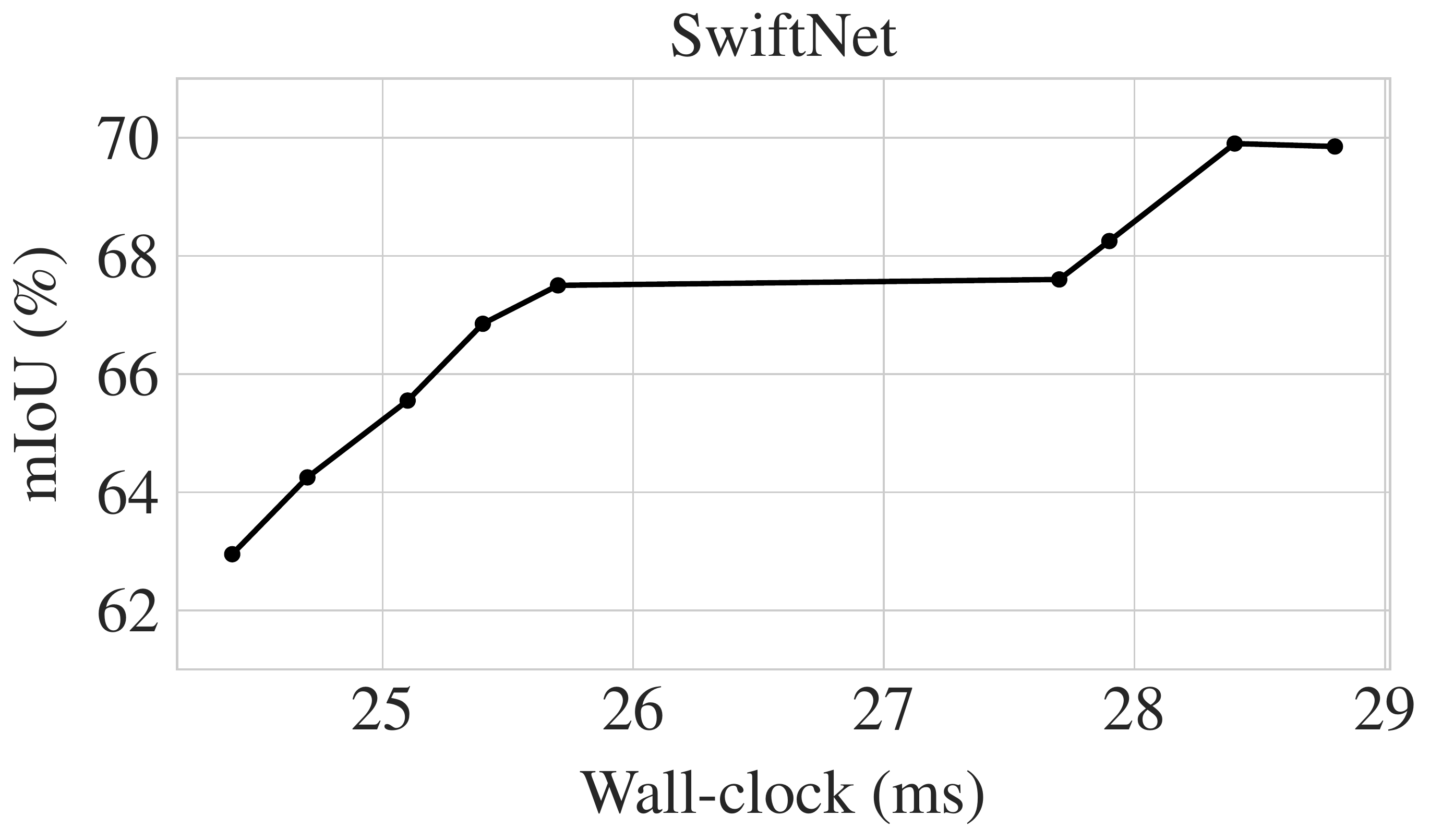}
  \caption{mIoU of CWM-SwiftNet trained with different bi-step generators.
  Each dot represents a CWM-SwiftNet model using a $\rho$-BG for $\rho$ going from $0.1$ to $0.9$ with steps of $0.1$.
  While \cref{fig:main} picks two values for $\rho$ and varies the width-multiplier $\alpha$, here we set $\alpha=1$ and vary $\rho$.
  We see that $\rho$ allows to \textit{modulate} the trade-off between speed and accuracy.
  }
  \label{fig:masks_tradeoff}
  \vskip -0.2in
\end{figure}

When using bi-step generators $\rho$-BG, the higher the value of $\rho$, the greater the number of channels processed at each time step and the longer the processing time.
In \cref{fig:masks_tradeoff}, we plot the mIoU of CWM-SwiftNet with different generators $\rho$-BG, for $\rho \in \{0.1, 0.2, 0.3, 0.4, 0.5, 0.6, 0.7, 0.8, 0.9\}$.
We use a fixed training offset $j=9$.

This plot highlights that the parameter $\rho$ allows controlling the trade-off between speed and accuracy, which is a desired and practical feature of our CWM models.

\subsection{Evaluation Offset}
\label{res:eval_offset}

For evaluating our model performance, we use Asymptotic Behavior Testing (ABT), as explained in \cref{exp:eval}.
We remind that ABT consists in feeding the model with inputs from $T - k$ to $T - 1$ before evaluating.
We additionally average the mIoU obtained with two consecutive $k$.

\begin{figure}
  \centering
  \includegraphics[width=0.9\linewidth]{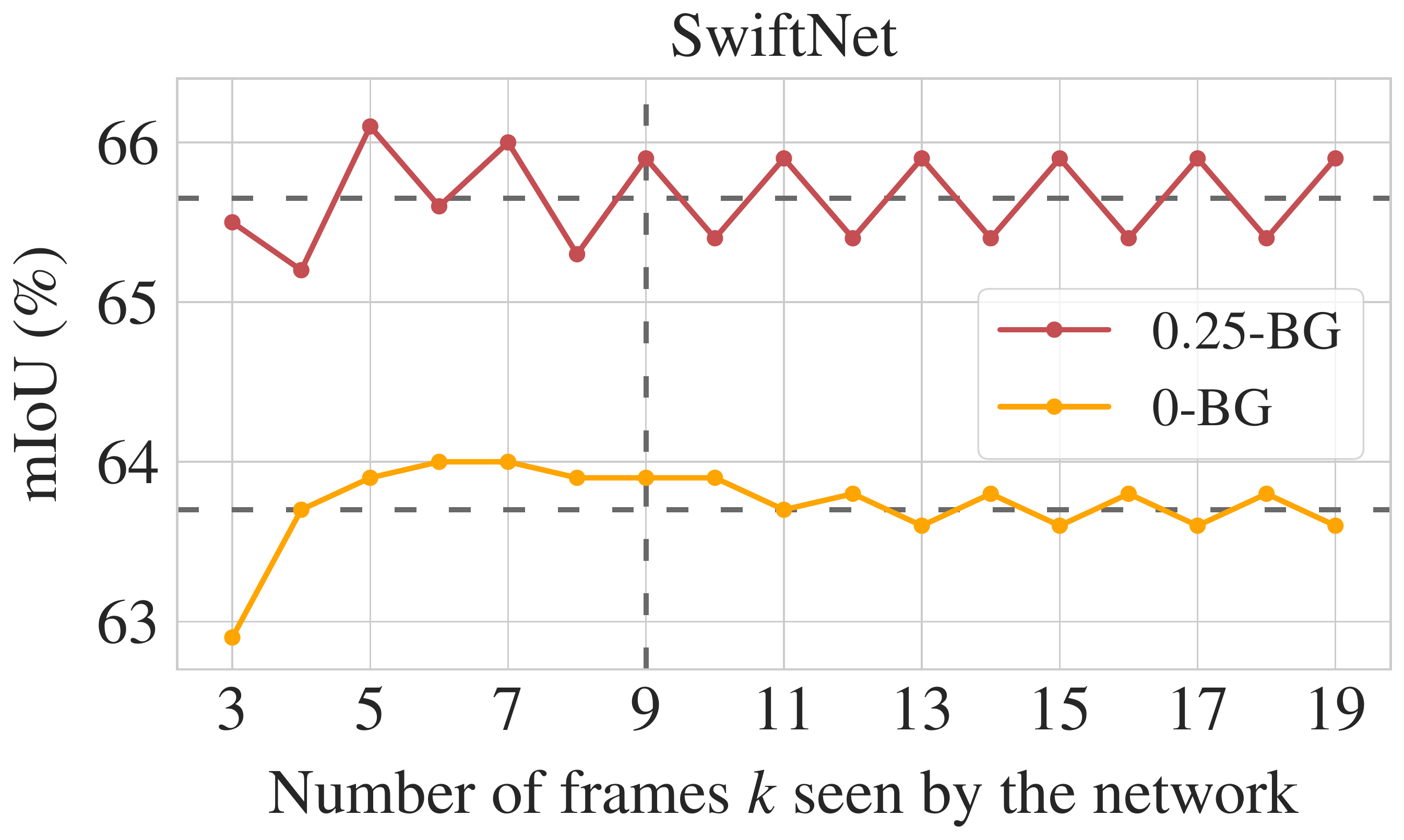}
  \caption{mIoU of CWM-SwiftNet as a function of the number of frames $k$ the network has seen.
  Horizontal dashed lines are values taken as the final mIoU of each network.
  The vertical dashed line marks the number of frames that networks have been trained with.
  We see that steady-state is reached after about 9 frames for $0.25$-BG and 12 frames for $0$-BG.
  The oscillation for higher $k$ is due to the two alternating masks of bi-step generators.
  }
  \label{fig:eval_offset}
  \vskip -0.1in
\end{figure}

\Cref{fig:eval_offset} plots our CWM-SwiftNet's mIoU depending on the number of frames $k$ it has processed, for all $k \in [3 \mathrel{.\,.} 19]$, and confirms the relevance of ABT and the averaging.

Indeed, for low $k$ values, the model is not in steady-state, and as $k$ increases, the model performance stabilizes, which is a strong argument for ABT.
In addition, the oscillation observed for higher $k$ is coherent with the use of bi-step generators that creates two masks used every other step and validates the need for averaging the mIoU computed with two consecutive $k$.

\subsection{Training Offset}
\label{res:train_offset}

In CWM models, the output prediction depends on the current input frame and previous computations, which is why we have introduced ABT, an evaluation procedure that evaluates the model's steady-state performance.
Therefore, it seems relevant to train the network when it operates in steady-state. To do so, we feed the model with $j$ inputs from $T-j$ to $T-1$, as explained in \cref{exp:training}.

\begin{figure}
  \centering
  \includegraphics[width=0.9\linewidth]{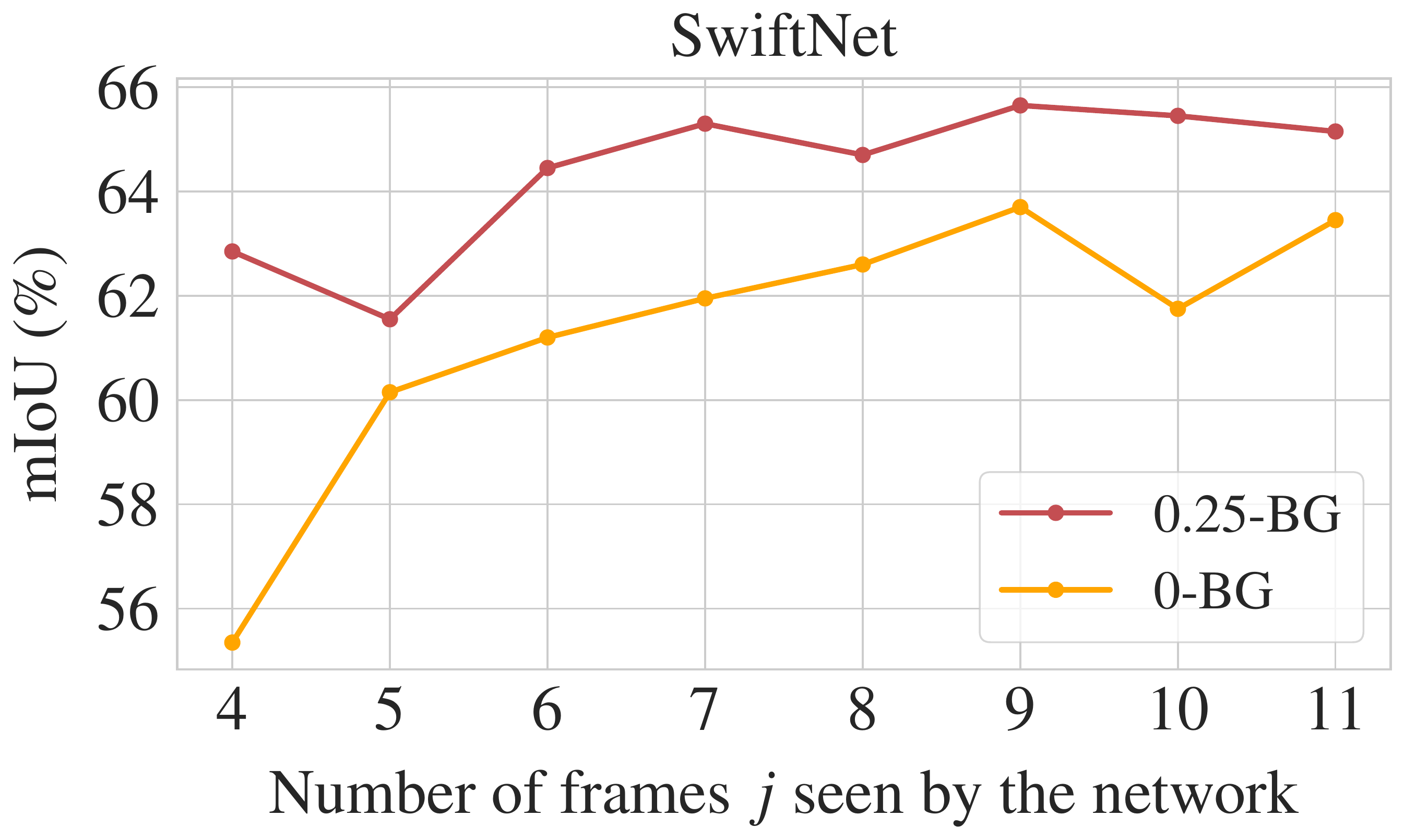}
  \caption{mIoU of a CWM-SwiftNet as a function of the length $j$ of the sequence it has been trained with. The mIoU appears to plateau from $j=7$, and $j=9$ gives the highest values.}
  \label{fig:train_offset}
\end{figure}

In this experiment, we plot the mIoU reached by models as a function of the number of frames $j$ used in training. The models used are 2 CWM-SwiftNet models with different mask generators $0$-BG and $0.25$-BG. The results in \cref{fig:train_offset} suggest that good steady-state performances start around $j=7$ and are best around $j=9$, which is why we used this last value in our main experiments.
However, we restrain from using a higher $j$, which increases training time and does not bring higher mIoU.
Note that the optimal $j$ value for training may be slightly different for other datasets and models, but that one can expect a value of this order of magnitude to reach steady-state.

\subsection{Bi-sequence training}
\label{res:mixing}

All experiments so far have used the ``bi-sequence training'' presented in \ref{exp:training}.
\Cref{tab:mixing} explores other multi-sequence training setups with SwiftNet.
In particular, it studies the single sequence training where a sequence is only used once with starting offset $T-j$, with $j=7$ in this experiment.
We also trained using $3$ and $4$ sequences.

The results are shown in \cref{tab:mixing}, and it appears that bi-sequence training is better than single sequence training ($+2.5\%$). Moreover, using additional offset sequences does not seem to improve the final mIoU, which motivates our choice of bi-sequence training for our other experiments.

\begin{table}
  \centering
  \begin{tabular}{l|cc}
    \toprule
    \multicolumn{1}{c|}{Starting offset of sequences} & mIoU (\%) \\
    \midrule
    $T$ -- $7 $ & 62.8 \\
    $ T$ -- $7 ,\; T$ -- $6 $ & \textbf{65.3} \\
    $ T$ -- $7,\; T$ -- $6,\; T$ -- $5 $ & 64.1 \\
    $ T$ -- $7,\; T$ -- $6,\; T$ -- $5,\; T$ -- $4$ & 64.3 \\
    \bottomrule
  \end{tabular}
  \caption{mIoU of a CWM-SwiftNet $0.25$-BG with different multi-sequence setup.
  The first column represents starting offset of sequences.
  For instance, ``$T-7, T-6$'' is the usual bi-sequence training, where one input sequence is used twice to optimize the model.
  It is used first as the subsequence $ [T-7 \;\dots\; T-1] $, then as the subsequence $ [T-6 \;\dots\; T-1] $.
  We see that bi-sequence is better than single sequence, and that using more sequences does not help.
  }
  \label{tab:mixing}
  \vskip -0.15in
\end{table}

\subsection{Ablation on CWM convolutions position}
\label{res:conv_pos}

When altering a network to use our CWM convolutions, we specified in \cref{exp:networks} that we do not replace the very first convolution in the stem and those used in skip connections.

In \cref{tab:position}, we study the performance in configurations where we use CWM convolutions instead of normal ones in the stem layer or in skip connections. The model is a CWM-SwiftNet using a $0.25$-BG. This table confirms that we should use normal convolutions in the stem and skip connections as it brings substantial mIoU increase.
The mIoU drop is particularly important for the stem, which may be due to the fact that there is no residual connection there to allow the new input to flow unchanged to further layers.

\begin{table}
  \centering
  \begin{tabular}{cc|c}
    \toprule
    Stem & Skip & mIoU (\%) \\
    \midrule
    CWM & CWM & 63.3 \\
    CWM & Standard & 64 \\
    Standard & CWM  & 65 \\
    Standard & Standard & \textbf{65.3} \\
    \bottomrule
  \end{tabular}
  \caption{Ablation experiment on CWM convolutions in the stem and skip connections of SwiftNet. CWM indicates a CWM convolution, Standard indicates normal convolution. This confirms that we should use normal convolutions in the stem and skip connections.}
  \label{tab:position}
  \vskip -0.15in
\end{table}

\section{Related Works}
Semantic segmentation, just as the rest of the computer vision literature, was shaken to the core by the success of deep-learning.
The seminal work \cite{long2015fully} introduced fully convolutional networks for segmentation, and several important methods followed, proposing to use decoder networks \cite{ronneberger2015u, badrinarayanan2017segnet}, spatial pyramidal pooling \cite{he2015spatial,zhao_pyramid_2016}, dilated convolutions \cite{yu2017dilated, chen2017deeplab}, and current best performing models are now mainly transformer based networks \cite{xie2021segformer,yuan2021segmentation,yan2022lawin,wu2021fully,strudel2021segmenter}.


One of the main limitation with current deep segmentation methods is the long inference time, and numerous works have addressed this \cite{li2019dfanet,zhao_icnet_2017,paszke2016enet,orsic2019defense,yu_bisenet:_2018,yu2020bisenet,hu2020temporally,li2020semantic,gao2021rethink,hong2021deep,fan2021rethinking,chao2019hardnet,nirkin2021hyperseg}.
These improvements allow carefully designed networks to run real-time, but any real-time network should be built as a future prediction network \cite{courdier2020real}.

On the future segmentation forecasting task, different approaches exist. Direct semantic forecasting introduced by  \cite{luc2017predicting} directly predicts the future segmentation map from past ones \cite{suh2018future,bhattacharyya2018bayesian,rochan2018future,chen2019multi}.
Flow based forecasting \cite{patraucean2015spatio,zhu2017deep,gadde2017semantic,nilsson2018semantic, terwilliger2019recurrent} uses optical flow computed from past frames to warp past segmentation into future ones.
Feature level forecasting predicts future intermediate features from past ones \cite{saric2020warp,luc2018predicting,chiu2020segmenting}.
Our technique stands at the intersection of direct and feature level forecasting as it indirectly predicts future features.

One could note that our training with masked convolution kernels resemble that of slimmable networks \cite{yu2018slimmable,yu2019universally} which perform a different sort of channel-wise masking with varying contiguous masks, although these masks are not designed nor appropriate for future video segmentation.

Our idea also echoes the fast TDNet \cite{hu2020temporally} method. When TDNet uses a different small sub-network at each time-step and combines features extracted from several past inputs, our CWM models uses different small convolution masks at each time-step and replace part of the past features with new ones.

\section{Conclusion}
In this paper, we have tackled the problem of fast prediction of video segmentation through the use of a new convolutional layer, that relies on the frames' temporal coherence.
Our simple design of this layer simultaneously reduces computation and gives access to features computed in previous steps, which speeds up and improves future segmentation prediction.

Our proposed CWM layer is not restricted to segmentation networks, and can be applied to any task involving prediction from a correlated input sequence. Moreover, the $\rho$ parameter of bi-step generators $\rho$-BG allows modulation of the speed/accuracy trade-off.

Experimentally, our proposed CWM-models achieve the same mIoU while performing fewer FLOPs, and taking lesser wall clock time than the original models in most cases.
Better yet, un-slimmed CWM-models even achieve higher mIoU than the originals.


\bibliographystyle{IEEE/IEEEtran}
\bibliography{IEEE/IEEEabrv,bibliography.bib}

\end{document}